\pdfoutput=1

\documentclass[11pt]{article}

\usepackage[final]{acl}

\usepackage{times}
\usepackage{latexsym}

\usepackage[T1]{fontenc}

\usepackage[utf8]{inputenc}

\usepackage{microtype}

\usepackage{inconsolata}

\usepackage{graphicx}

\definecolor{nb}{HTML}{006EB8}
\usepackage{hyperref}
\hypersetup{
  colorlinks   = true,    
  urlcolor     = nb,    
  linkcolor    = nb,    
  citecolor    = nb      
}
\newcommand{\datasetname}{{\tt{XplainLLM}}}

\usepackage{booktabs} 
\usepackage{enumitem}
\usepackage{amsmath,bm,amssymb}
\usepackage[ruled,lined,linesnumbered]{algorithm2e}
\usepackage{multirow}
\usepackage{tcolorbox}

\definecolor{softGray}{HTML}{FFFFFF}
\definecolor{deepBlue}{HTML}{003366}
\definecolor{darkGray}{HTML}{333333}
\definecolor{highlightBackground}{HTML}{E0DBE8}
\newcommand{\highlight}[1]{\colorbox{highlightBackground}{#1}}
\newtcolorbox{stage1}{
    colback=softGray, 
    colframe=deepBlue, 
    coltext=darkGray, 
    title=Instruction for Generating Explanation,
    fontupper=\fontsize{8pt}{1pt}\selectfont,
}

\title{\datasetname{}: A Knowledge-Augmented Dataset for Reliable Grounded Explanations in LLMs}

\author{Zichen Chen\textsuperscript{1} \hfill Jianda Chen\textsuperscript{2}\hfill Ambuj K. Singh\textsuperscript{1}\hfill Misha Sra\textsuperscript{1}\\
\textsuperscript{1}University of California, Santa Barbara\\ 
  \textsuperscript{2}Nanyang Technological University, Singapore\\
  \texttt{\{zichen\_chen, ambuj, sra\}@ucsb.edu} \hfill \texttt{
jianda001@ntu.edu.sg}}

\begin{document}
\maketitle
\begin{abstract}
Large Language Models (LLMs) have achieved remarkable success in natural language tasks, yet understanding their reasoning processes remains a significant challenge. We address this by introducing \datasetname{}, a dataset accompanying an explanation framework designed to enhance LLM transparency and reliability.
Our dataset comprises 24,204 instances where each instance interprets the LLM's reasoning behavior using knowledge graphs (KGs) and graph attention networks (GAT), and includes explanations of LLMs such as the decoder-only Llama-3 and the encoder-only RoBERTa. 
\datasetname{} also features a framework for generating grounded explanations and the \textit{debugger-scores} for multidimensional quality analysis. Our explanations include \emph{why-choose} and \emph{why-not-choose} components, \textit{reason-elements}, and \textit{debugger-scores} that collectively illuminate the LLM's reasoning behavior.  
Our evaluations demonstrate \datasetname{}'s potential to reduce hallucinations and improve grounded explanation generation in LLMs. \datasetname{} is a resource for researchers and practitioners to build trust and verify the reliability of LLM outputs.
\end{abstract}

\section{Introduction}

As the capabilities and applications of large language models (LLMs) continue to expand~\citep{liu2023pre,achiam2023gpt,touvron2023llama,jiang2024mixtral}, the need for transparency and interpretability in their reasoning behavior has become increasingly urgent~\citep{arrieta2020explainable}. Traditional methods~\citep{ribeiro2016should,lundberg2017unified,casalicchio2019visualizing} allow us to get insights into the reasoning behind language model outputs, but they fall short of providing a complete picture, leaving the logic behind complex decision obscured~\citep{huang2023can}. This gap presents a significant barrier in applications where model decision transparency is important, such as healthcare~\citep{ghosh2024clipsyntel}, law~\citep{cheong2024not}, and public services~\citep{musumeci2024llm}.

Current methods for explaining LLM's reasoning behavior primarily focus on the analysis of parameter changes~\citep{clark2019does,jacovi2021contrastive,bills2023language} and chain-of-thought (CoT) based self-explanation~\citep{huanginner,li2023trustworthy}. Analysis of parameter changes bases the explanations on self-attention weights in models like BERT~\citep{kenton2019bert} and GPT-2~\citep{radford2019language}, deducing correlations between input tokens and the model's predictions. However, the relationships highlighted in these generated explanations are difficult to understand for humans. CoT-based self-explanation, on the other hand, iteratively generates rationales step-by-step. Due to the inherent constraints in LLMs, these explanations often have hallucinations and can not reflect the real reasoning process~\citep{huang2023can}.

\begin{table*}[htbp]
    \centering
    \resizebox{\textwidth}{!}{
    \begin{tabular}{@{}c|ccccccc@{}}
    \toprule
        Dataset & Size & Answer Format & Expl. Format & Source & Model Match? & Self-Explanatory? & "Why Not" Included? \\
        \midrule
        CoS-E & 9,500 & MC & NL & Human & $\times$ & $\times$ & $\times$ \\
        ECQA & 10,962 & MC & NL & Human & $\times$ & $\checkmark$ & $\times$ \\
        Neuron & 307,200 & Neuron & NL + Score & Model & $\checkmark$ & $\times$ & $\times$ \\
        \datasetname{} & 24,204 & MC & NL & Model & $\checkmark$ & $\checkmark$ & $\checkmark$ \\
    \bottomrule
    \end{tabular}
    }
    \caption{Comparison of prevalent explanation datasets with \datasetname{}, detailing instance count (Size), answer types (Answer Format: e.g., multiple-choice (MC)), explanation styles (Explanation Format: e.g., natural language (NL)), origin (Source), alignment with model reasoning (Model Match?), necessity of human intervention to deduce the reasoning (Self-Explanatory?), and inclusion of reasons for alternative answer rejection ("Why Not" Included?).}
    \label{tb:dataset}
\end{table*}

We introduce \datasetname{}, a dataset accompanying an explanation framework designed to enhance transparency, explainability, and understandability in LLM reasoning behaviors. By integrating knowledge graphs (KGs) and Graph Attention Networks (GAT)~\citep{veličković2018graph}, we construct a structured and reliable dataset that anchors explanations in reasoning-relevant knowledge. 
We link the LLM reasoning process to the entities and relations within KGs to help provide an intuitive and interpretable representation of the LLM's decision-making process. 
Our process also helps facilitate model tuning, debugging, robustness evaluation and demonstration in in-context learning. 
\datasetname{} provides a structured explanation of two distinct types of LLMs: Llama-3-8B~\citep{touvron2023llama} (decoder-only model) and RoBERTa-large~\citep{liu2019roberta} (encoder-only model). A total of 24,204 instances are included in the dataset. 
The explanations are tied to two models' reasoning processes, derived from their performance on the CommonsenseQA~\citep{talmor2019commonsenseqa} challenge.

Additionally, we introduce an explanation framework that utilizes a retrieval-based method to support generating grounded explanations for LLMs. This framework operates without the need for additional model training, utilizing \datasetname{} as a knowledge base to retrieve the most relevant data points to the given query. The selected data points serve as demonstration examples for in-context learning~\citep{dong2022survey}, enabling the LLMs to generate explanations that are more grounded in the reasoning process.

We evaluate the quality of the explanations in \datasetname{} through human and automated evaluations. The overall quality of explanations achieves an average score of 0.87/1.00 by human evaluators, and an average of 0.89/1.00 by automated evaluators. We evaluate our framework by comparing the performance of LLMs with and without our framework, and the results show that LLMs under our framework outperform the benchmark, with a performance gap extending to 20\%. We also evaluate the quality of the explanations generated by our framework, and the results underscore the quality of our explanations on multiple metrics.

In summary, we make two key contributions to the field of explainable AI for LLMs: (1) an explanation dataset of model reasoning behavior, and (2) a framework for improving the interpretability of LLMs through structured, grounded explanations. To the best of our knowledge, \datasetname{} is the first dataset to provide structured and grounded explanations for LLM reasoning behavior.

\section{Related Work}

\paragraph{Interpretability in LLMs}

Explainable AI aims to address the issue of interpreting the outcomes of language models \citep{li2023trustworthy, wiegreffe2021measuring, madsen2022post}. One of its goals is to generate explanations that enable humans to easily understand the decision-making process. 
\citet{zelikman2022star,zhang2023towards,wang2023can} utilize gradual strategies that iteratively generates the rationales step-by-step. \citet{huanginner,chen2023teaching,tanneru2024quantifying,chakraborty2023ranking} utilize the CoT to find the rationale and apply the reasoning capabilities of LLMs to domain tasks. 
However, these explanations are inherently constrained in capturing prompt-specific reasoning, which often generates hallucinations and can not reflect the real reasoning of LLMs~\citep{turpin2024language}.

Another goal is focused on explaining in a trustworthy way. \citet{rajani-etal-2019-explain} introduce an explainable factor to minimize the risk of unreasonable explanation generation. \citet{chen-etal-2021-kace} integrate the external knowledge to generate why and why-not counterfactual explanations. 
\citet{zelikman2022star} apply self-checker mechanism to ensure trusted rationals. 
However, these method fail to accurately capture the core reasoning of LLMs. In contrast, our work enhances LLM trustworthiness and deepens human understanding of its reasoning behavior, improving the potential in end-user applications.

\paragraph{Explanation Datasets}
The explainable datasets for language models can be categorized into three types~\citep{wiegreffe2021teach}: (1) highlights: provide input elements such as words and phrases, as explanations to a predicted output~\citep{NEURIPS2018_4c7a167b,deyoung-etal-2020-eraser,yin-etal-2021-context, bills2023language}; (2) free-text explanations: provide readable textual explanations in words or sentences~\citep{rajani2019explain, sap-etal-2020-social,Brahman_Shwartz_Rudinger_Choi_2021}; (3) structured explanations: provide natural language explanation but are constrained by the explanation writing process \citep{aggarwal2021explanations,jhamtani-clark-2020-learning,inoue-etal-2020-r4c}. 
Different from these, our explanation incorporates highlighted reason-elements and guided instruction to generate a free-text explanation. Our explanation is structured and grounded in the reasoning process, enhancing the trustworthiness and comprehensiveness of the content.
We present a comparison with prevalent explanation datasets~\citep{rajani2019explain, aggarwal2021explanations, bills2023language} in Table \ref{tb:dataset}.

\section{\datasetname{}: Dataset, Explanation Framework and Debugger-Score}\label{sec:dataset}

\datasetname{} serves three essential purposes in interpreting LLMs' reasoning behavior. First, it utilizes KG and GAT to interpret LLM through parameter changes, collecting these explanations to build a dataset. The LLMs we used are Llama-3-8B~\citep{touvron2023llama} (decoder-only) and RoBERTa-large~\citep{liu2019roberta} (encoder-only). Second, we provide an explanation framework for generating faithfully grounded explanations without additional training.
Third, we introduce the \textit{debugger-score}, which is designed for multidimensional analysis to quantify the quality of explanations, supporting our framework for comprehensively evaluating and improving LLM explainability.

\subsection{Task Definition and Collection Method}\label{sec:method}
\begin{figure}[t!]
    \centering
    \includegraphics[width=0.99\columnwidth]{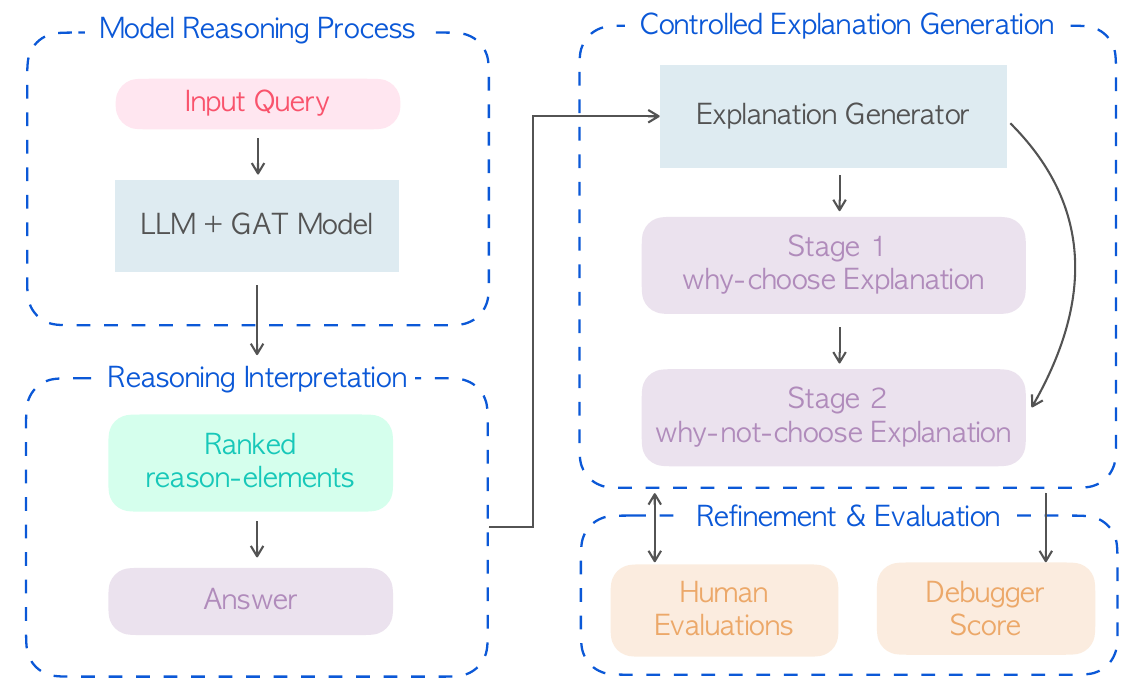}
    \caption{Overview of \datasetname{} in LLM Reasoning Interpretation and Explanation Generation.}
    \label{fig:overview}
\end{figure}
The primary goal of \datasetname{} is to enhance the interpretability of LLMs through grounded explanations.
We define the task as generating explanations that clarify the decision-making processes behind model predictions. We use QA tasks to generate instances for our dataset. The overview of the collection process is shown in Figure \ref{fig:overview}. A more detailed data collection description is shown in Appendix~\ref{sec:collection}.

The LLM's reasoning is grounded in a structured KG, which is used to identify the most salient features that influence the model's predictions. We employ GAT to analyze the KG's structure and identify the influence of specific nodes and edges that are salient to the model's decision-making process.
Each instance in \datasetname{} is formulated as follows:
\begin{equation}
    \text{Instance} = \left((Q,A), \text{Explanation}\right)
\end{equation}
where $(Q,A)$ is the question-answer pair and $\text{Explanation}$ includes:
\begin{itemize}[itemsep=0.1mm,leftmargin=*]
    \item A \emph{why-choose} explanation, detailing the reason behind the model's answer choice.
    \item A \emph{why-not-choose} explanation, detailing reasons against alternative choices.
    \item \textit{Ranked reason-elements}, identified through GATs that analyze the KG's structure to identify critical influencing elements.
    \item A \textit{debugger-score} for each explanation, quantifying its faithfulness, completeness, accuracy and overall quality.
\end{itemize}

\paragraph{Graph-Based Reasoning Interpretation.}
To produce the aforementioned explanation, we introduce a graph-based interpreting method to learn the features that influence the model's decision-making process.
We first extract the key elements from the KG $g$. 
This process involves identifying nodes and edges within the $g$ that are relevant to the input question and answer pair. We incorporate node relevance scores into this retrieval process, using the LLM's knowledge to guide the pruning of the $g$:
\begin{equation}
    G_e = \text{PruneKG}(Q, A, g, s_i)
\end{equation}
where $s_i$ represents the relevance score for each node $i$ in the retrieved graph, calculated using LLM's probability function that assesses the alignment of node embeddings with the input context $(Q, A)$. The function $\text{PruneKG}$ evaluates the semantic relationship between node embeddings and the query. This extraction leverages the LLM's knowledge to focus on the most informative elements for the given QA context. The algorithm for constructing the $G_e$ is provided in Appendix~\ref{sec:graph}.

Once the relevant subgraph $G_e$ is obtained, we use a GAT to determine the significance of each node and edge in contributing to the model's output. Each node \( i \) at \( k \)-th layer is represented by a feature vector \( h_{ki} \).
The attention $\alpha_{ij}$ for each node pair $(i, j)$ are computed using a softmax function over a parameterized self-attention mechanism $a$ that captures the relationship dynamics:
\begin{equation}
    \alpha_{ij} = \frac{\exp(a(h_{ki}, h_{kj}))}{\sum_{l \in \mathcal{N}(i)} \exp(a(h_{ki}, h_{kl}))}
\end{equation}
where $\mathcal{N}(i)$ denotes the neighbors of node $i$. 


The updated node features $h_{k+1,i}$ are computed by aggregating the features of neighboring nodes weighted by their respective attention scores:
\begin{equation}
    h_{k+1,i} = \sigma\left(\sum_{j \in \mathcal{N}(i)} \alpha_{ij} W f_m(h_{kj}, u_i, r_{ij}) \right) + h_{ki}
\end{equation}
where $f_m$ is a multi-layer perceptron (MLP) that processes features of neighboring nodes considering their types and interrelations. $W$ is a weight matrix, $\sigma$ is a non-linear activation function. We provide the details of the GAT model in Appendix~\ref{sec:graph_algo}.

We define the probability of selecting an answer \(v\) from the set \(A\) by leveraging both the representation embeddings from the language model (\( \mathbf{H}^{LM} \)) and the graph-based reasoning features (\( h_K \) and \( \alpha_K \)) extracted from our subgraph \(G_e\):
\begin{equation}
    P(a|q) \propto \exp(\text{MLP}(\mathbf{H}^{LM}, h_{K}, \alpha_K))
\end{equation}
where \( h_K \) represents the output features from the final layer of our $K$-layer graph reasoning network, and \( \alpha_K \) are the attention coefficients. To this end, we map the LLM's reasoning to the graph features. The extracted attention features are mapped to their corresponding nodes in the \(G_e\), and we select the top \(n\) nodes with the highest attention scores for generating the explanations.

\paragraph{Controlled Explanation Generation.}
Upon obtaining the reasoning features, we transform them into structured and human-understandable explanations through a two-stage instructional process. The top \(n\) nodes are selected as the key \emph{reason-elements} set \( R \), which guides the explanation generator model \( \mathbb{F} \) to construct the explanations. The explanation generation process includes: (1) \emph{why-choose} explanation: the reasoning behavior behind the model's choice, and (2) \emph{why-not-choose} explanation: the rationale for dismissing other potential answers.
The instruction for \emph{why-choose} stage is: 
``$Basis:[TASK\_TYPE], Input:[Q,A], Output[y',R], Explanation (Stage 1):[y'/y]$''. The output of stage 1 named $E_{why}$ is used as the input for stage 2.
The instruction for \emph{why-not-choose} stage is: ``\(Explanation (Stage 2):[E_{why}, \ A \setminus {y'}]\)''. The details of the instruction are provided in the Appendix~\ref{sec:instruction}.

\subsection{Explanation Framework for Grounded Explanations}
\begin{figure}
    \centering
    \includegraphics[width=0.99\columnwidth]{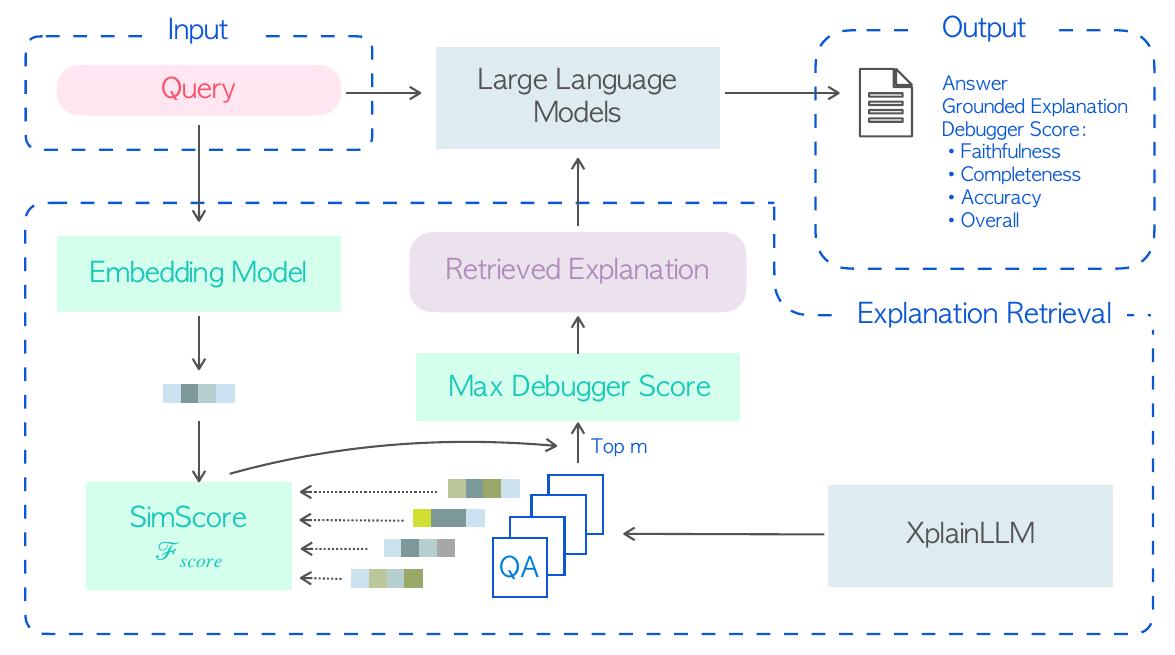}
    \caption{Explanation Framework for Grounded Explanation Generation in LLMs.}
    \label{fig:framework}
\end{figure}
To enhance the usability of \datasetname{} and facilitate the generation of grounded explanations for different types of LLMs (especially for private LLMs, e.g., GPT-4), we introduce an explanation framework that leverages the collected dataset to generate faithfully grounded explanations without additional model training. The framework is illustrated in Figure \ref{fig:framework}.
The process is divided into three steps:

\paragraph{Embedding Calculation.} 
When receiving a new query \((Q_{\text{new}}, A_{\text{new}})\), its embeddings \(\mathbf{e}_{QA_{\text{new}}}\) is calculated using the an embedding models. 
To generalize our framework, we use voyage-2-large model from VOYAGE AI~\footnote{\url{https://docs.voyageai.com/docs/embeddings}}, as our embedding model to extract the embeddings, due to its state-of-the-art performance in generalist text embedding~\footnote{\url{https://huggingface.co/spaces/mteb/leaderboard}}.

\paragraph{Similarity Computation and Retrieval.}  We retrieve the most contextually relevant instances by computing the cosine similarity between new query embedding \(\mathbf{e}_{QA_{new}}\) and instance embedding \(\mathbf{e}_{QA}\) in \datasetname{} $\mathcal{E}$:
\[
\mathcal{F}_{\text{score}}(\mathbf{e}_{QA_{new}}, \mathbf{e}_{QA}) = \frac{\mathbf{e}_{QA_{new}}^\top \mathbf{e}_{QA}}{\|\mathbf{e}_{QA_{new}}\|_2 \|\mathbf{e}_{QA}\|_2}
\]
This function $\mathcal{F}_{\text{score}}$ scores each instance \(\text{sim}(\mathbf{e}_{QA_{new}}, \mathbf{e}_{QA})\) for relevance. The \(\mathbf{e}_{QA}\) can be pre-computed and stored in the dataset for efficient retrieval.
To accelerate the retrieval process, we provide embeddings for each instance in \datasetname{}, using voyage-2-large.

\paragraph{Instance Selection and Explanation Generation.}
The top \( m \) instances with the highest similarity scores, \(\text{sim}(\mathbf{e}_{QA_{new}}, \mathbf{e}_{QA})\), are selected. Each instance may contain multiple explanations from different LLMs, denoted as \(\mathcal{E}_t\), where \( t \) indexes the instances. For each instance, we select the explanation \( e^* \) that maximizes the \textit{debugger-score} set $D$:
    \[
    e^* = \arg\max_{e \in \mathcal{E}_t} \sum_{d \in D} w_d \cdot D(e, d)
    \]
where and \( w_d \) are the weights reflecting user preferences for each dimension. This selection is influenced by user-specified preferences which dictate the importance of various dimensions of explanation quality, such as faithfulness or accuracy. We will introduce the \textit{debugger-score} in Section \ref{sec:debugger}.
These selected instances are used as in-context learning examples for targeted LLM to generate grounded explanations.

\subsection{Debugger-Score for Explanation Analysis}\label{sec:debugger}

To improve the understanding of generated explanations, we introduce the \emph{debugger-score} to evaluate the quality of explanations. Inspired by the method of transformer debugging~\citep{bills2023language}, our \textit{debugger-score} simulates a ``perfect'' LLM to benchmark against the actual LLM's reasoning. It quantifies the quality of explanations by assessing:
\begin{enumerate}[itemsep=0.1mm,leftmargin=*]
    \item \textbf{Faithfulness:} How accurately the explanations reflect the actual reasoning of the LLM.
    \item \textbf{Completeness:} Whether the explanations cover all essential aspects of the reasoning process.
    \item \textbf{Accuracy:} The correctness of the explanation in terms of factual and contextual relevance.
    \item \textbf{Overall:} The overall quality of the explanation, combining the above dimensions.
\end{enumerate}

The \textit{debugger-score} utilizes predefined instructions to guide the evaluation, focusing on identifying discrepancies between the simulated ``perfect'' LLM and the actual LLM. Our evaluation method quantifies the quality of explanations, providing a measure of where the LLM's reasoning succeeds or falls short. Our \textit{debugger-score} is used to enhance the reliability and transparency of the explanations. Further details on the implementation and functionality of the \textit{debugger-score} can be found in the Appendix~\ref{sec:app-debugger}.

\section{Dataset Overview and Preparation}

\subsection{Dataset Description}

\paragraph{Schema.} \datasetname{} contains fields that correspond to the QA pair, the model's predicted answer, the ground-truth label, and an explanation set.

\paragraph{Explanations Set.} The explanation set includes a set of 50 \emph{reason-elements}, e.g., words or phrases, sorted by attentions, a set of top-5 \emph{reason-elements}, a \emph{why-choose} explanation in free-text form, a \emph{why-not-choose} explanation also in free-text form. An example instance is shown in Appendix~\ref{sec:app-example}.

\paragraph{Statistics.} \datasetname{} includes 24,204 instances of explanations, split according to the official CommonsenseQA's partitioning into three sets: the training, development (dev), and testing sets. The average word count of $E_{why}$ and $E_{why-not}$ are 94.77 and 85.74 respectively, resulting in an aggregate count of approximately 180.81 words per whole explanation. A more detailed breakdown of the average word count is provided in Table \ref{tab:statis}. Additional statistics can be found in Appendix~\ref{sec:statistics}.

\begin{table}[t]
\centering
\resizebox{\columnwidth}{!}{%
\begin{tabular}{@{}c|ccc@{}}
\toprule
             & Why-choose & Why-not-choose & Whole Explanation \\ \midrule
Overall      &  94.77     & 85.74          & 180.81            \\
Training Set &  94.41     & 85.22          & 179.63            \\
Dev Set      & 93.00      & 84.54          & 178.44            \\
Testing Set  & 96.89      & 87.46          & 184.35            \\ \bottomrule
\end{tabular}%
}
\caption{The average word counts of \emph{why-choose} explanation, \emph{why-not-choose}  explanation and whole explanation in our \datasetname{} dataset.}
\vspace{-12pt}
\label{tab:statis}
\end{table}

\subsection{Data Preparation}

\datasetname{} captures and analyzes the reasoning behavior of LLMs on CommonsenseQA dataset~\citep{talmor2019commonsenseqa}. 
CommonsenseQA serves as a foundational benchmark for assessing the commonsense reasoning capabilities of these models. 

We select Llama-3-8B and RoBERTa-large as LLMs for our dataset as they exemplify decoder-only and encoder-only LLMs respectively, providing a comprehensive view of different model architectures in language understanding.
The models are fine-tuned on CommonsenseQA's official training set, to understand and interpret the complexities of commonsense reasoning. We utilize ConceptNet \citep{speer2017conceptnet} as our KG to obtain $g_e$. This KG captures commonsense concepts and their interrelations. We use a 5-layer GAT model to extract the reasoning paths.
We use GPT-3.5-turbo~\citep{ouyang2022training} and GPT-4-turbo~\citep{achiam2023gpt} as explanation generator model $\mathbb{F}$ to generate a natural language explanation in a sentence or a paragraph. To ensure the quality of our dataset, we conduct a post-generation evaluation. All explanations undergo human review. Human evaluators identify inaccuracies, and any discrepancies in explanations, and return to $\mathbb{F}$ for refinement. This procedure mitigates potential issues from model-generated explanations, guaranteeing clarity and relevance aligned with human understanding. We also provide embeddings of the $(Q, A)$ pair for each instance in the dataset. The embeddings are generated using the voyage-large-2. The \textit{debugger-score} is calculated using GPT-4-turbo.
Further experiment specifics and data collection procedures are provided in the Appendix~\ref{sec:app-exp} and \ref{sec:collection}.

\section{Experiments and Evaluation}\label{sec:eval}

\subsection{Evaluation Methodology}
We evaluate \datasetname{} and explanation framework through two main perspectives:
\begin{enumerate}[leftmargin=*]
    \item \textbf{Explanation Quality Evaluation:} The quality of the explanations generated by the LLMs is assessed via a dual approach:
    \textbf{(1) Human Evaluation} - Experts and crowdsourcing review the explanations, and \textbf{(2) Automated Evaluation} - GPTs evaluate the explanations.

    \item \textbf{Framework Effectiveness:} We measure the impact of our proposed methods on the groundedness of newly generated explanations and the performance of the LLMs.
    This includes:
    \textbf{(1) Grounded Explanation Assessment} - Using the \textit{debugger-score} to evaluate how well the explanations are grounded in factual content, and
    \textbf{(2) Performance Analysis} - We evaluate changes in the accuracy of the LLM outputs by comparing metrics before and after applying our framework.
\end{enumerate}
Specifically, the evaluation metrics for explanation quality assessment are human-centered metrics, following the guidelines of~\citet{hoffman2018metrics}. Each explanation is assessed using seven evaluative questions that explore different aspects of the explanation's impact and quality. The metrics encompass overall quality, understandability, trustworthiness, satisfaction, detail sufficiency, completeness, and accuracy. Evaluators allocate scores to these questions using a three-point Likert scale: 1 (disagree), 2 (neutral), and 3 (agree). Subsequently, scores are normalized to the range [0, 1]. Higher scores suggest better quality. 
Detailed definitions are provided in the Appendix~\ref{sec:human}. 

\subsection{Explanation Quality Evaluation}
We conducted human and automated evaluations to go beyond the technical evaluation of the explanations.
The human evaluation involved three experts with NLP backgrounds and 50 general users via Prolific\footnote{https://www.prolific.com}.
Our participant pool was gender-balanced, and comprised of native English speakers with at least a high school education.
Experts and users rate 20 randomly selected explanations based on guidelines adapted from \citep{hoffman2018metrics} to ensure consistency and mitigate bias. Automated evaluations are performed using GPT-3.5-turbo and GPT-4 to parallel human judgment, quantifying performance with standardized scores. Detailed methodologies and participant instructions are provided in Appendix~\ref{sec:question}.

\begin{table}[t]
    \centering
    \resizebox{\columnwidth}{!}{
    \begin{tabular}{@{}c|ccc@{}}
    \toprule
        & Expert <-> GPT-3.5 & Expert <-> GPT-4 & GPT-3.5 <-> GPT-4 \\ \midrule
    \ \ $\rho $  & 0.70           & 0.60         & 0.66        \\ \bottomrule
    \end{tabular}
    }
    \caption{Correlation coefficient ($\rho$) between overall quality scores evaluated by expert, GPT-3.5 and GPT-4.}
    \label{tab:expert-corr}
\end{table}

\paragraph{Results of Expert and Automated Evaluation.}
The feedback from human experts highlighted the distinctiveness of our explanations compared to existing methods. One expert remarked,
``\textit{In comparison to prior explanations, these explanations provide a more intuitive understanding of the LLM's reasoning behavior. The explanations are cogent, and even in instances of erroneous predictions, the underlying reasoning remains transparent and comprehensible.}''
This feedback underscores the clarity and transparency of our explanations. 

\begin{table*}[t]
    \resizebox{\textwidth}{!}{
    \begin{tabular}{@{}c|ccccccc@{}}
    \toprule
     & Overall Quality & Understandability & Trustworthiness & Satisfaction & Sufficiency of detail & Completeness & Accuracy \\ \midrule
    GPT-3.5      & 0.98 & 0.98 & 0.98 & 0.98 & 0.98 & 0.98 & 0.98 \\
    GPT-4        & 0.90 & 0.93 & 0.87 & 0.87 & 0.88 & 0.87 & 0.88 \\
    Human Expert & 0.91 & 0.97 & 0.95 & 0.89 & 0.98 & 0.97 & 0.93 \\
    Crowdsourcing & 0.85 & 0.89 & 0.86 & 0.80 & 0.83 & 0.81 & 0.85 \\\bottomrule
    \end{tabular}
    }
    \caption{Evaluation by automated evaluator GPT-3.5, GPT-4, human experts and crowdsourcing, on 7 evaluation metrics.}
    \label{tab:simulator_score}
\end{table*}

\begin{figure}
    \centering
    \includegraphics[width=0.4\textwidth]{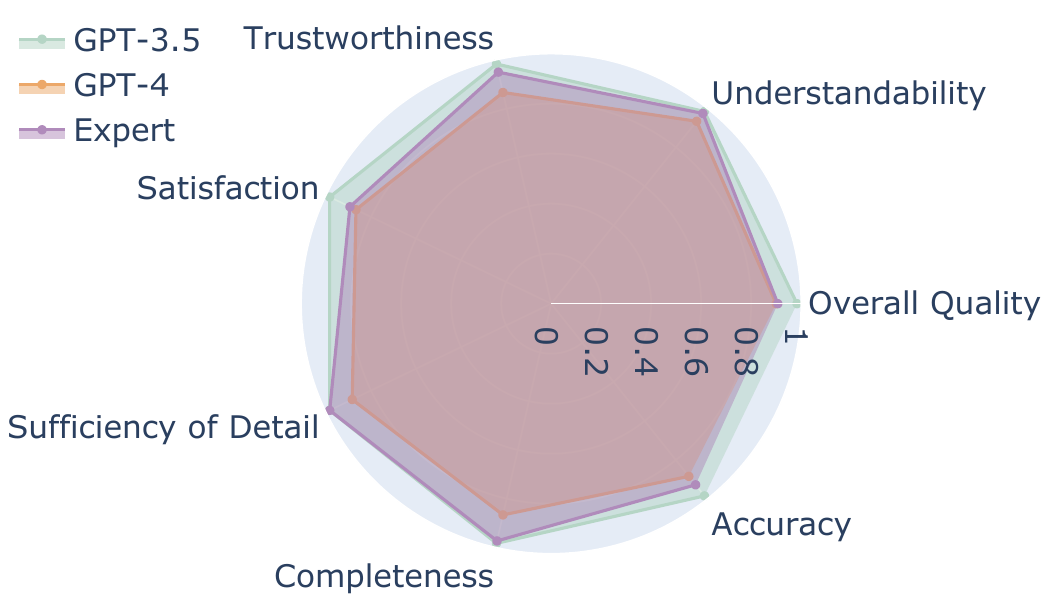}
    \caption{Evaluation by human experts, automated evaluator GPT-3.5 and GPT-4. } \label{fig:simulator}
\end{figure}

The results are summarized in Table \ref{tab:simulator_score}. Human experts assign an average score of 0.93/1.00 across seven evaluation metrics, with ``understandability'' and ``completeness'' receiving the highest scores. The automated evaluators, GPT-3.5 and GPT-4, assign average scores of 0.91/1.00 and 0.92/1.00, respectively. 
The performance of these automated evaluators aligns closely with human expert evaluations across dimensions, as shown in Figure \ref{fig:simulator}. 

Further insights into the human-like understanding of automated evaluators and their assessment of explanations are detailed in Table \ref{tab:expert-corr}. This data shows a significant agreement between the automated evaluators and human experts. Such findings further support the credibility and value of our explanations.

\begin{figure}[t]
    \centering
    \includegraphics[width=0.4\textwidth]{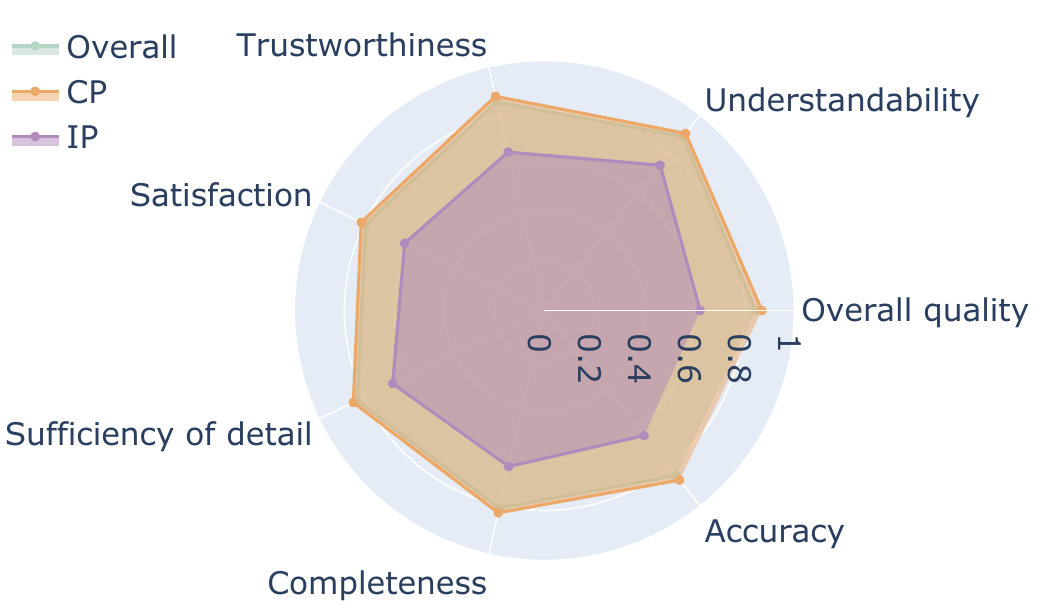}
    \caption{Human evaluation of explanations: Overall, CP, and IP. Note that the CP scores align closely with the overall scores.}
    \label{fig:overall}
    \vspace{-5pt}
\end{figure}

\paragraph{Results of Crowdsourcing Evaluation.}

we present the average scores from crowdsourcing on eight metrics, as depicted in Figure \ref{fig:overall}. These scores reflect evaluations of the overall explanations, as well as separate assessments for explanations of correct predictions (CP) and incorrect predictions (IP). The details of our analysis are discussed below.

Participants assigned a high average score of 0.87/1.00 to the overall quality of our explanations, indicating a favourable perception and underscoring their above-average clarity. 
The explanations received an average understandability score of 0.89/1.00, demonstrating their clarity. The low variance of 0.26 suggests consistent comprehension among participants. However, a detailed analysis shows a disparity based on the LLM's prediction accuracy: explanations for correct predictions (CP) were highly rated at 0.91/1.00 with a variance of 0.26, while explanations for incorrect predictions (IP) scored lower at 0.74/1.00 with a variance of 0.65, indicating less clarity and greater variability in participant responses. 

In terms of trustworthiness, our explanations scored an average of 0.88/1.00 for CP. A Pearson correlation coefficient of 0.71 between trustworthiness and understandability confirms a strong positive relationship, suggesting that clearer explanations enhance participants' trust in the LLM's outputs.

Overall satisfaction with our explanations is high, with 86\% of participants stating that the explanations meet or exceed their expectations. 97.36\% of the explanations are considered sufficiently detailed. 
The completeness of our explanations also received high marks, with an average score of 0.81/1.00 and a median score of 1.00/1.00, suggesting that over half of the participants find the explanations to be entirely comprehensive. However, the distribution may reflect differences in the evaluators' familiarity with AI or occasional oversimplifications by the model. The accuracy of the explanations are rated at 0.84/1.00, with a noticeable disparity between CP at 0.87/1.00 and IP at 0.64/1.00, highlighting how the LLM's prediction accuracy significantly influences the perceived accuracy of explanations. Furthermore, a Pearson correlation of 0.68 between accuracy and trustworthiness indicates that more accurate explanations are considered more trustworthy.

The positive feedback from our crowdsourcing evaluations robustly validates \datasetname{}, demonstrating its effectiveness in conveying the complexities of the LLM's decision-making in a clear, trustworthy, and satisfying manner to users.

\subsection{Framework Evaluation}

\begin{table}[t!]
    \centering
    \resizebox{\columnwidth}{!}{
    \begin{tabular}{l|c|c|cccc}
    \toprule
    \textbf{Model} & \textbf{\#P} & \textbf{Version} & \textbf{Faithfulness} & \textbf{Completeness} & \textbf{Accuracy} & \textbf{Overall} \\
    \midrule
    \multirow{2}{*}{\textbf{gpt-3.5-turbo}} & \multirow{2}{*}{Unknown} & Vanilla & \textbf{3.50} & 2.95 & \textbf{3.65} & \textbf{3.37} \\
                                             &  & \datasetname{} & 3.45 & \textbf{3.05} & 3.55 & 3.35 \\
    \midrule
    \multirow{2}{*}{\textbf{gpt-4-turbo}} & \multirow{2}{*}{Unknown} & Vanilla & 3.67 & 3.10 & 3.95 & 3.57 \\
                                          &  & \datasetname{} & \textbf{4.05} & \textbf{3.65} & \textbf{4.10} & \textbf{3.93} \\
    \midrule
    \multirow{2}{*}{\textbf{llama3-8b}} & \multirow{2}{*}{8.02B} & Vanilla & \textbf{3.50} & 2.65 & \textbf{3.60} & \textbf{3.25} \\
                                        &  & \datasetname{} & 3.35 & \textbf{2.85} & 3.35 & 3.18 \\
    \midrule
    \multirow{2}{*}{\textbf{llama3-70b}} & \multirow{2}{*}{70.6B} & Vanilla & 3.6 & 2.85 & 3.80 & 3.42 \\
                                         &  & \datasetname{} & \textbf{3.95} & \textbf{3.10} & \textbf{4.05} & \textbf{3.70} \\
    \midrule
    \multirow{2}{*}{\textbf{mixtral-8x7b}} & \multirow{2}{*}{46.7B} & Vanilla & 3.70 & 2.90 & 3.80 & 3.47 \\
                                           &  & \datasetname{} & \textbf{3.75} & \textbf{3.05} & \textbf{3.80} & \textbf{3.53} \\
    \bottomrule

    \end{tabular}}
    \caption{Comparison of Vanilla and \datasetname{} Versions of Models with \textit{debugger-score}.}

    \label{table:debugger}
\end{table}

\begin{figure}
    \centering
    \includegraphics[width=0.4\textwidth]{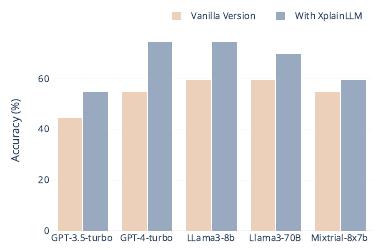}
    \caption{Accuracy comparison of vanilla version and with \datasetname{} version for different models.}
    \label{fig:acc}
\end{figure}

In evaluating our proposed framework, we include five LLMs: GPT-3.5-turbo~\citep{brown2020language}, GPT-4-turbo~\citep{achiam2023gpt}, Llama3-8B~\citep{touvron2023llama}, Llama3-70B~\citep{touvron2023llama}, and Mixtral-8x7B~\citep{jiang2024mixtral}. We compare the vanilla versions of these models with the versions enhanced by \datasetname{}. The results are summarized in Table \ref{table:debugger}.
We specifically selected 20 questions from \datasetname{} designed to challenge models by exposing their tendency to produce hallucinations. This choice is based on the need to test the framework's ability to ground model's explanation. 
We then evaluate five LLMs both with and without the enhancements provided by \datasetname{}, allowing us to explore how our framework performs across different scales and architectures.
The benchmarks for this evaluation are focused on four key metrics: faithfulness, completeness, accuracy, and overall performance, as shown in Table \ref{table:debugger}. We further quantified the impact of our framework by comparing the accuracy rates of the vanilla version to those enhanced with our modifications, as detailed in Figure \ref{fig:acc}. 

Our results show that performance variations across different model architectures and configurations, as demonstrated in Table \ref{table:debugger}. Notably, the GPT-4-turbo model, when enhanced with our framework, demonstrates exceptional performance across key metrics. It scores 4.05/5.00 in Faithfulness, 3.65/5.00 in Completeness, and 4.10/5.00 in Accuracy, culminating in an Overall score of 3.93/5.00. These high scores suggest that our framework not only improves the overall output quality but also ensures that the LLM's reasoning is grounded in faithful knowledge, thus enhancing both the clarity and reliability of the model's behavior explanation.

We also observe a consistent improvement in accuracy across different LLMs when our framework is applied, as shown in Figure \ref{fig:acc}, which implies a scalable utility of our framework. 
We find the GPT-4-turbo model exhibits the most significant improvement. This may suggest that our enhancements are effective in assisting more complex LLMs to ground their reasoning in faithful knowledge, thereby reducing hallucinations and improving interpretability.

By comparing the detailed reasoning explanation of the models with and without our framework, we observe that the explanations generated under the vanilla version tend to generate outputs that are not entirely supported by input data (hallucinations). In contrast, the explanations generated under the \datasetname{} version are more grounded in factual content, and exhibit greater faithfulness. 

We find our framework can guide the LLMs toward a more grounded and data-driven approach in generating outputs. This is helpful for applications where precision and reliability are paramount, such as in legal, medical, or safety-critical environments. Furthermore, the consistent improvements across LLMs of varying capabilities suggest that our framework is robust and scalable, capable of enhancing a wide range of AI systems. This broad applicability suggests potential for widespread adoption in enhancing the transparency and accountability of AI decision-making processes.

\section{Conclusion}

We introduce \datasetname{}: a knowledge-augmented dataset paired with an explanation framework designed to enhance the interpretability of LLMs. Our dataset and framework provide a way for LLMs to generate reliable and grounded explanations without additional training. Through the use of \textit{debugger-score}, we provide a multidimensional analysis of quantitatively evaluate the quality of explanations. Our evaluations demonstrate that \datasetname{} not only grounds explanations in reasoning behavior, but also helps LLMs reduce hallucinations and improve their performance. The dataset and code are available at \url{https://github.com/chen-zichen/XplainLLM_dataset.git}. We release them under the MIT license to encourage further research in explainable AI.

\section*{Limitation}
Committed to transparency and rigorous analysis, we acknowledge potential limitations in our dataset. Since our \textit{reason-elements} $R$ is originally derived from $g_e$, any inherent limitations or inaccuracies within used KG could influence the quality of our explanations.  

\section*{Ethical Considerations}
While \datasetname{} and its accompanying explanation framework provides advancements in the transparency and accountability of LLMs, several risks might exist.
First, the reliance on KGs and structured data may lead to biases embedded in these sources, potentially skewing the explanations. Secondly, incorrect knowledge augmentation could mislead users about the accuracy of the explanations. Additionally, there is a risk that users might over-rely on the \textit{debugger-score} without critical assessment, potentially overlooking context-specific inaccuracies. It is essential for future work to continuously refine \datasetname{}, address detected biases, and enhance the robustness of the framework to mitigate these risks.

\bibliography{custom}

\newpage
\appendix

\section{Graph Construction Algorithm} \label{sec:graph}

\begin{algorithm}
\caption{Sub-graph Construction (PruneKG)}
\KwData{Graph \(g\) with nodes \(n\), input content \(QA\), encoding function of LLM \(f_{enc}\), MLP \(f_{s}^{node}\), Number of top nodes to select \(N\)}
\KwResult{Pruned graph \(g_e\)}
\Begin{
    Initialize an empty list \( \textit{node\_scores} \) \;
    \For{each node \( n \) in \( g \)}{
        Obtain the embedding of $n$: \( \mathcal{B} \leftarrow f_{enc}(n||QA) \) \;
        Compute the relevance score of $n$: \( s_i \leftarrow \textit{sigmoid}(f_{s}^{node}(\mathcal{B})) \) \;
        Append \( (n, s_i) \) to \( \textit{node\_scores} \) \;
    }
    Sort \( \textit{node\_scores} \) in descending order based on \( s_i \) \;
    Select the top \( N \) nodes from the \( \textit{node\_scores} \) list \;
    Create a new graph \( g_e \) with the selected \( L \) nodes, preserving their edges and properties \;
    \Return \( g_e \) \;
}
\end{algorithm}

\section{Details of Graph Attention Network} 
\label{sec:graph_algo}
In section \ref{sec:method}, we detail the method for interpreting the LLM's reasoning behavior through graph-based techniques. We provide supplementary calculations and algorithmic details in this section.

We describe the process for updating the node features in a graph using a GAT in Equation (4). Here, each node \( i \) updates its feature vector \( h_{k+1,i} \) based on the features of its neighboring nodes \( N(i) \). \( f_m \) is transformation function, modeled as a MLP, that maps the input features \( h_{kj} \), \( u_i \) , and \( r_{ij} \) into a higher-dimensional space. specifically, \( u_i \) is the one-hot vector encoding the type of node \( i \), and \( r_{ij} \) is the relation embedding denoting the relationship type between nodes \( i \) and \( j \), calculated by:
\begin{equation}
    r_{ij} = f_\theta (i, u_{ij}) = f_\theta (i, u_{i} \parallel u_{j}),
\end{equation}
where $u_{ij}$ is an one-hot vector encoding the type of connection between nodes $i$ and $j$, and $u_{ij}$ is the concatenation of $i$ and $j$.

\section{Instruction for Explanation Generation}\label{sec:instruction}

Due to the space constraints, we provide detailed guidelines and instructions for generating explanations in this section.

\textbf{Basis:} Given a LM augmented with a graph attention network to extract key reasoning elements for decision-making. 
The task is \highlight{[TASK\_TYPE]}.

\textbf{Input:} The question is: \highlight{[$Q$]}. The Answer Options are: \highlight{[$A$]}

\textbf{Output:} The model predicted choice \highlight{[$y'$]}. Based on the Ranked Reason-elements: \highlight{[$\mathcal{R}$]}

\textbf{Explanation (Stage 1):} 
Explain the LM's reasoning process for selecting \highlight{[$y'$]} over the other options. Provide concise explanations for why each reason-element supports \highlight{[$y'$]} as the predicted choice. Focus on the LM's behavior and the significance of the Ranked Reason-elements. Your response should be short and concise.

\textbf{Explanation (Stage 2):} Based on the \highlight{[$E_{why}$]}, explain why this LM makes the other options less likely \highlight{[$A\setminus\{y'\}$]}. Your response should be short and concise.

\section{Details of Debugger-Score} \label{sec:app-debugger}

The \textit{debugger-score} is a metric that quantifies the quality of the explanations generated by the LLMs. The score evaluates explanations based on multiple dimensions such as faithfulness, accuracy, and completeness. By measuring how well the explanations align with a ``perfect'' targeted LLM, the debugger score provides a comprehensive evaluation of the generated explanations. This metrics is useful for ensuring that the explanations are not only plausible but also grounded in facts, enhancing trust of explanations generated by LLMs. This instruction assesses explanations based on three dimensions: faithfulness, completeness, and accuracy. 

\subsection{Instructions for Debugger-score Calculation}

\textbf{Prompt System:} Evaluators, assuming the role of LM debuggers with expertise in model parameter changes, assess explanations from the perspective of how model parameters influence decision-making. The assessment focuses on whether the explanation accurately reflects the computational and statistical mechanisms utilized by the LM.

\textbf{Prompt Content:} Evaluators are presented with a task where the LM is augmented with key reasoning elements derived from its operation. This includes the question, answer options, the LM's prediction, and the corresponding explanation.

\textbf{Evaluation Criteria:}
\begin{itemize}
    \item \textbf{Faithfulness:} Does the explanation accurately represent the underlying computational processes and data-driven mechanisms used by the LM to reach its conclusion?
    \item \textbf{Completeness:} Does the explanation encompass all significant computational strategies and data insights relied upon by the LM to make the decision?
    \item \textbf{Accuracy:} How precisely does the explanation reflect the true capabilities and decision-making processes of the LM, considering its design, training data, and functional algorithms?
\end{itemize}

\textbf{Scoring:} Evaluators are instructed to score each dimension on a scale from 1 to 5, where 1 indicates the lowest level of adherence (poor) and 5 indicates the highest (excellent). The scoring guide emphasizes balanced evaluation, advising against overly strict judgments. 

\section{Experiments}\label{sec:app-exp}
In this section, we describe the details of our evaluation that were omitted in Section~\ref{sec:eval} due to space constraints.

\subsection{Model Parameters} \label{sec:params}
To train our GNN, we use a dropout rate of 0.2, a batch size of 64, and a learning rate of 1e-5, optimized with RAdam. The model is fine-tuned on a single NVIDIA A100 GPU for approximately 3 hours. 
Our KG containing 799,273 nodes and 2,487,810 edges. Our $g_e$ is pruned based on KG to retain 200 high-ranking nodes with a hop size of 2. 
The GNN, specifically, consists of 200 dimensions and 5 layers. The learning rate in our experiments is 1e-3.

\section{Detailed Data Collection}\label{sec:collection}

\begin{figure}[h]
  \centering
  \includegraphics[width=0.99\columnwidth]{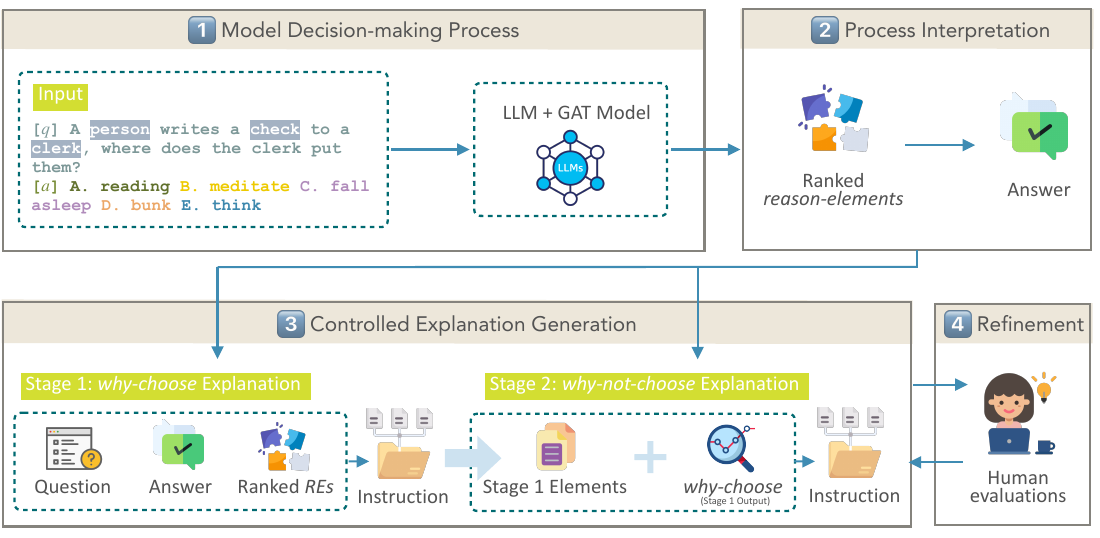}
  \caption{Data Collection Process.
  }
  \label{fig:process}
\end{figure}

Figure \ref{fig:process} shows the process of data collection:

    (1) Given a question, we retrieve its relevant knowledge using the KG. The retrieved graph is then pruned based on scores influenced by the LLM, resulting in what we term the \textit{element-graph}. The \textit{element-graph} is processed by a specialized GAT model (known formally as Decision Interpretation). Leveraging attention mechanisms, we obtain the essential representations for interpretation.\\
    (2) The model's decision-making is interpreted through the ranked \textit{reason-elements} and the predicted answer.\\
    (3) A controllable dual-stage process generates the explanations: Stage 1: The initial phase focuses on generating the ``why-choose" explanations. Stage 2: Building upon the outputs and elements of Stage 1, we then generate the ``why-not-choose" explanations. \\
    (4) We conduct a human evaluation to identify errors in the explanations. If discrepancies arise, explanations are reverted to Step 3 for refinement. This process not only helps prevent potential issues arising from bad explanations generated by the LLM but also maintains human-aligned clarity and relevance.

\section{Instance Example} \label{sec:app-example}

We present an example from our dataset to illustrate the data instance.

\textbf{Question:} John carred for Lucy but had trouble expressing it.  Lucy was disturbed by John's inability to express affection and felt that he was what?

\textbf{Answers:} A. being mean, B. negligence, C. disinterest, D. misunderstood, E. unfeeling

\textbf{Label:} unfeeling

\textbf{Predicted Label Index:} 4

\textbf{Label Matched:} true

\textbf{Reason-elements:} "enraged", "delay", "abiogenesis", "sneerer", "helpable", "begrudge", "mollify" ...

\textbf{Top-5 Reason-elements:} "enraged", "delay", "abiogenesis", "sneerer", "helpable"

\textbf{Explanation (Why):} The model selected ``unfeeling'' primarily due to how it processed the emotional descriptors in the scenario and its understanding of the term definitions. Though the key reason-elements like 'enraged', 'delay', 'abiogenesis', 'sneerer', and 'helpable' seem unrelated at first, they could connect to various aspects of emotional disconnect and misunderstanding. 'Enraged' hints at strong emotional responses, albeit negative, which possibly led the model to associate less expressed emotions to being ``unfeeling''. The term 'sneerer', indicating derision, might be linked by the model to John's perceived dispassion. The word 'helpable' suggests a situation needing improvement, reflecting on Lucy's feelings towards John's emotional expression deficits.

\textbf{Explanation (Why-Not):} 1. ``being mean'': The context didn't explicitly describe John's behavior as intentionally harmful or spiteful, which 'being mean' would imply. The reason-elements don't directly associate with malicious intent. 2. ``negligence'': Although John's behavior could be interpreted as neglectful, the elements like 'enraged' do not support an outright disregard, but rather an emotional complexity. 3. ``disinterest'': This choice could somewhat fit, but the model likely found stronger connections in emotional response terms suggesting not just a lack of interest but a lack of feeling. 4. ``misunderstood'': While the context and reason-elements like 'sneerer' might suggest misunderstandings, 'unfeeling' directly refers to a perceived absence of emotion which seemed more fitting to Lucy's disturbance.

\textbf{debugger-score:} Faithfulness: 4 | Completeness: 3 | Accuracy: 4

The format of our dataset is as follows:
\begin{tcolorbox}[colback=white, sharp corners, boxrule=1pt, title=\textbf{Data Schema}]
\textbf{question:} typeof(string)

\textbf{answers:} typeof(list\_of\_strings)

\textbf{label:} typeof(string)

\textbf{predicted\_label:} typeof(string)

\textbf{label\_matched:} typeof(boolean)

\textbf{concept:} typeof(list\_of\_strings)

\textbf{topk:} typeof(list\_of\_strings)

\textbf{explanation\_why:} typeof(string)

\textbf{explanation\_why\_not:} typeof(string)

\textbf{debugger\_score:} typeof(string)

\textbf{embedding:} typeof(list\_of\_floats)
\end{tcolorbox}

\section{Explanation Statistics} \label{sec:statistics}
\begin{figure}[h]
  \centering
  \begin{minipage}[t]{0.47\textwidth}
    \includegraphics[width=1\textwidth]{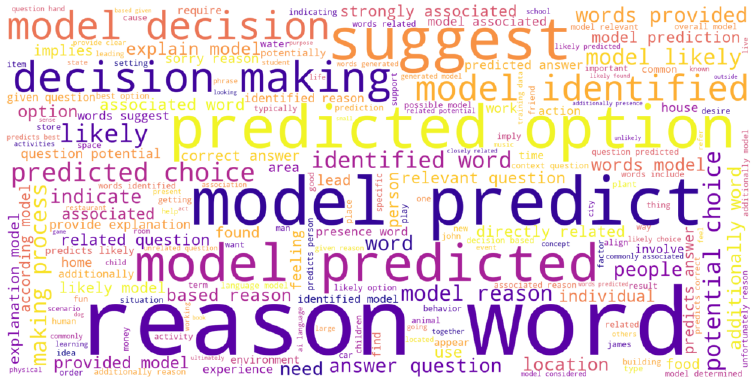}
  \caption{\emph{why-choose} explanations.}
  \label{fig:cloud-why}
  \end{minipage}
  \begin{minipage}[t]{0.47\textwidth}
      \includegraphics[width=1\textwidth]{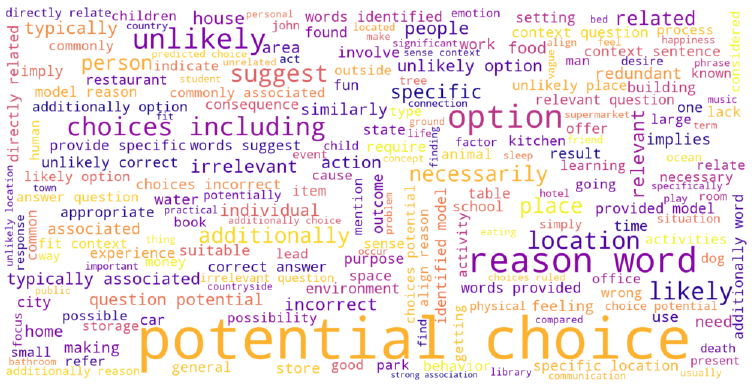}
      \caption{\emph{why-not-choose} explanations.}
  \label{fig:cloud-why-not}
  \end{minipage}
\end{figure}

Figure \ref{fig:cloud-why} is a word cloud showing the most frequently appearing words in the \emph{why-choose} explanations. From this figure, we have a clear indication that \emph{why-choose} explanations focus on explaining, comprehension, and interpreting predictions made by the target model.

Figure \ref{fig:cloud-why-not} presents a word cloud for \emph{why-not-choose} explanations. We note that these explanations outline the reasons behind the non-selection of specific options as predicted answers. Furthermore, \emph{why-not-choose} explanations emphasize how the target model determines the likelihood of different answer choices. We also observe that the target model handles a wide array of topics, which can be crucial components in the ``why not'' reasoning process. 
\section{Evaluation Materials} \label{sec:material}

\subsection{Questions and Evaluation Instructions} \label{sec:question}

For each instance, we include a set of question, answer choices, model prediction, and explanation. To evaluate the quality of the explanation, we provide seven questions for evaluators. 
Each question includes three score levels: 1 for disagree, 2 for neutral, and 3 for agree. The questions and instructions in our evaluation are as follows: 

\textbf{Q0: This is a good explanation}

1. \textbf{Disagree}: The explanation is illogical or inconsistent with the question and/or does not adequately cover the answer choices.

2. \textbf{Neutral}: The explanation is somewhat logical and consistent with the question but might miss some aspects of the answer choices.

3. \textbf{Agree}: The explanation is logical, consistent with the question, and adequately covers the answer choices.

\textbf{Q1: I understand this explanation of how the AI model works.}

1. \textbf{Disagree}: The explanation is unclear or contains overly complex terms or convoluted sentences.

2. \textbf{Neutral}: The explanation is somewhat understandable but might contain complex terms or convoluted sentences.

3. \textbf{Agree}: The explanation is clear, concise, and easy to understand.

\textbf{Q2: I trust this explanation of how the AI model works.}

1. \textbf{Disagree}: The explanation is unclear or contains overly complex terms or convoluted sentences.

2. \textbf{Neutral}: The explanation is somewhat credible but contains some elements that I find doubtful or questionable.

3. \textbf{Agree}: The explanation is credible and aligns with my understanding of how AI models work.

\textbf{Q3: This explanation of how the AI model works is satisfying.}

1. \textbf{Disagree}: The explanation does not meet my expectations and leaves many questions unanswered.

2. \textbf{Neutral}: The explanation somewhat meets my expectations but leaves some questions unanswered.

3. \textbf{Agree}: The explanation meets my expectations and satisfies my query.

\textbf{Q4: This explanation of how the AI model works has sufficient detail.}

1. \textbf{Disagree}: The explanation lacks detail and does not adequately cover the AI model's decision-making.

2. \textbf{Neutral}: The explanation provides some detail but lacks thoroughness in covering the AI model's decision-making.

3. \textbf{Agree}: The explanation is thorough and covers all aspects of the AI model's decision-making.

\textbf{Q5: This explanation of how the AI model works seems complete.}

1. \textbf{Disagree}: The explanation does not adequately cover the answer choices and leaves many aspects unexplained.

2. \textbf{Neutral}: The explanation covers most answer choices but leaves some aspects unexplained.

3. \textbf{Agree}: The explanation covers all answer choices and leaves no aspect unexplained.

\textbf{Q6: This explanation of how the AI model works is accurate.}

1. \textbf{Disagree}: The explanation does not accurately reflect the AI model's decision-making.

2. \textbf{Neutral}: The explanation somewhat reflects the AI model's decision-making but contains some inaccuracies.

3. \textbf{Agree}: The explanation accurately reflects the AI model's decision-making.

\subsection{Human-centered Metrics for Explanation Quality Evaluation} \label{sec:human}
The meaning of metrics used in the human-centered evaluation are as follows:
\begin{enumerate}[leftmargin=*]
    \item
    \textbf{Overall quality} reflects the overall effectiveness of explainability.
    It reveals how effectively explanations convey the decision-making process of the AI models to the human users. 
    \item 
    \textbf{Understandability} evaluates how well a human can comprehend the model's output and explanations. It captures the clarity and coherence of the generated text. 
    \item 
    \textbf{Trustworthiness} measures the human evaluator's confidence in the model's outputs and explanations. It evaluates whether the explanations appear reliable, credible, and based on sound reasoning. 
    \item 
    \textbf{Satisfaction} captures the overall contentment of the evaluator with the explanations. It measures whether the outputs meet the evaluator's needs and expectations in terms of quality, relevance, and utility.
    \item 
    \textbf{Sufficiency of detail }evaluates whether the explanations provide a sufficient level of detail. It evaluates whether the responses are adequately descriptive and provide all necessary information to fully answer the question or task.
    \item 
    \textbf{Completeness} measures whether the explanations address the decision behaviors of the model. 
    \item 
    While we also measure \textbf{accuracy} objectively, the human evaluation of accuracy assesses whether the explanations align with the evaluator's knowledge or expectations. It measures whether the explanations can reflect if the model's outputs are factually correct and contextually appropriate.

\end{enumerate}

\end{document}